\documentclass{article}
\usepackage{spconf,amsmath,graphicx,subcaption}
\usepackage{hyperref}

\title{Resolving referring expressions in images with labeled elements}

\name{Nevan Wichers, Dilek Hakkani-T\"ur, Jindong Chen}
\address{Google AI, Mountain View, CA, USA\\
{\tt wichersn@google.com, dilek@ieee.org, jdchen@google.com} }

\begin{document}
\ninept
\maketitle
\begin{abstract}
Images may have elements containing text and a bounding box associated with them, for example, text identified via optical character recognition on a computer screen image, or a natural image with labeled objects. We present an end-to-end trainable architecture to incorporate the information from these elements and the image to segment/identify the part of the image a natural language expression is referring to. We calculate an embedding for each element and then project it onto the corresponding location (i.e., the associated bounding box) of the image feature map. We show that this architecture gives an improvement in resolving referring expressions, over only using the image, and other methods that incorporate the element information. We demonstrate experimental results on the referring expression datasets based on COCO, and on a webpage image referring expression dataset that we developed.
\end{abstract}
\section{Introduction}
\begin{figure}[h]
{\rm {\bf User:} Show me something funny, maybe cat videos?}\\
{\rm {\bf Agent:} Here are funny cat videos from YouTube}\\
\begin{center} 
\includegraphics[width=8.5cm]{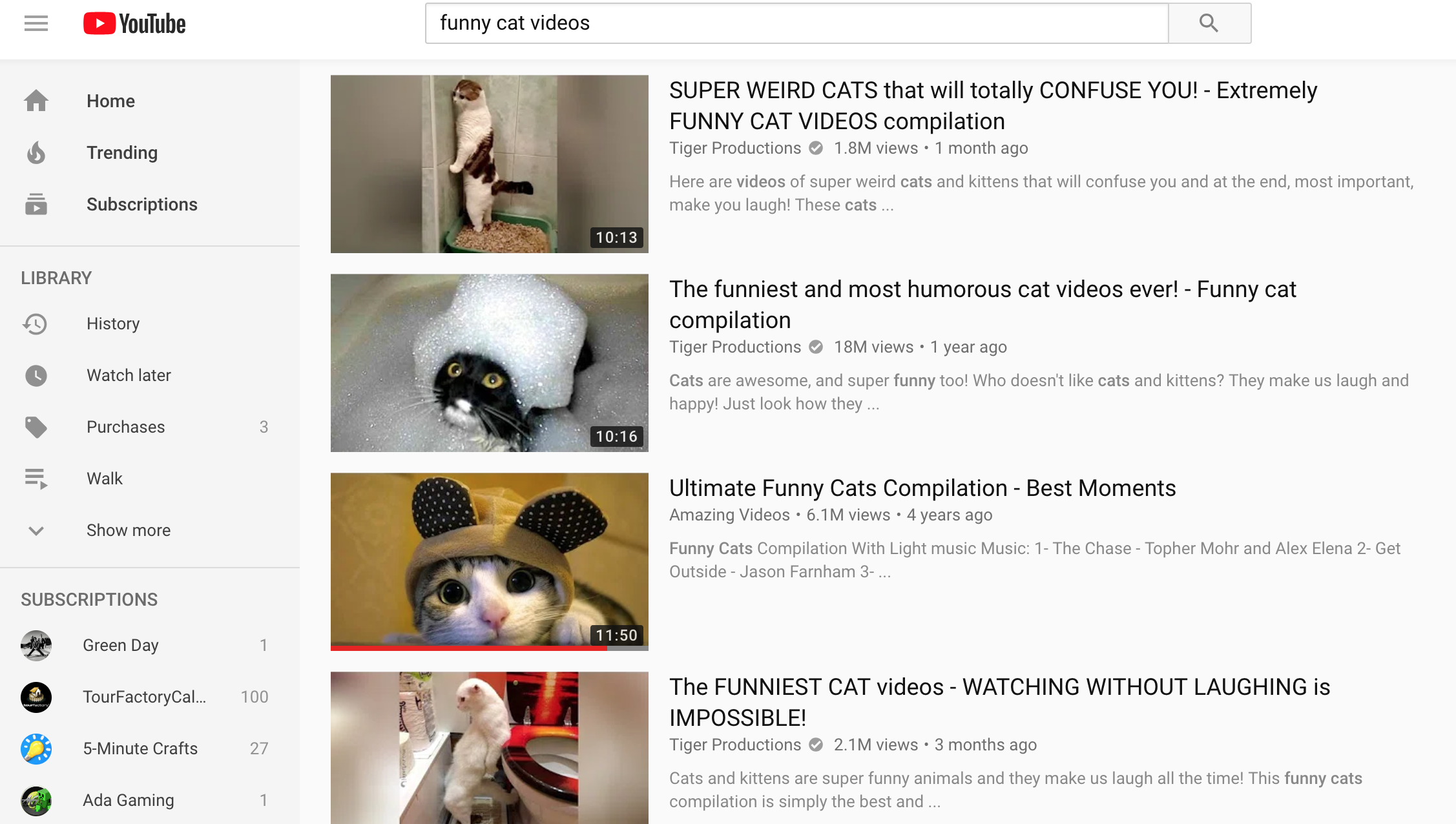}
\end{center}
{\rm {\bf User:} I wanna watch the {\em super weird cats}.}
\caption{Example visual conversation where the user refers to the first video on the screen by referencing the text on the page. This example shows how our model fits into a dialogue system.}
\label{fig:conversationExample}
\end{figure}

Referring expression resolution on images is a well researched problem. In this task, given an image and a natural language referring expression, the goal is to identify the part of the image being referred to by the expression. Our motivation for this work is to enable visual dialogues between a user and an agent where both have access to a shared screen image. In this setting, the image often has elements associated with it that could be candidates for referring expressions. By elements we mean annotations of an area on an image with text and bounding boxes which correspond to the image. For example, in a conversational system, the agent might present the user with a list of movies where each movie is an element with associated text and a bounding box. Users may naturally refer to these visual items in their utterances in the following turns via referring expressions. Another use case is a user navigating their computer hands free by specifying where to click, see Figure \ref{fig:conversationExample} for an example interaction. Another use case is a visually impaired user interacting with the real world through an agent that takes snapshots of the world and allows the user to ask questions about objects in the environment using referring expressions.

In this context, a user may refer to objects in a variety of ways, such as by their position, or by their relationship to another element. For visual interfaces, we think a very common scenario will be a user referring to an object by the text in a screen element. Theoretically, a model which takes only the image and referring expression into account could solve this by learning to do optical character recognition (OCR) based on its training data. However, this will require a large amount of data, so practically a model which uses the screen elements explicitly will help. For natural images, associated text is not always available, but an object recognition system could produce such annotations. Similar to the visual interfaces, we expect users will commonly be referring to the objects by including descriptive text generated by an image segmentation or object recognition system in the referring expressions.

\begin{figure}[h]
\includegraphics[width=.5\textwidth]{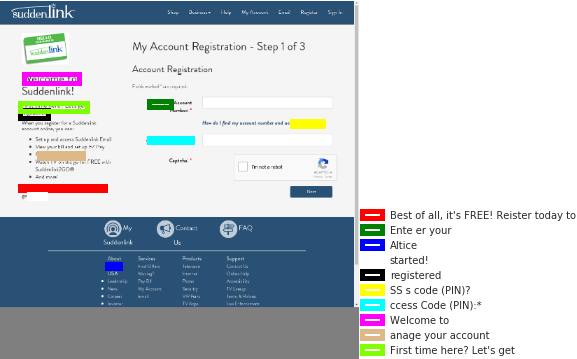}
\includegraphics[width=.4\textwidth]{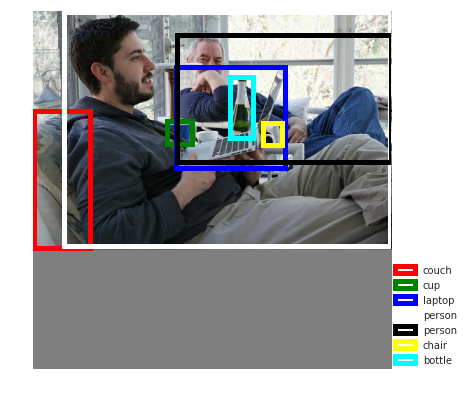}
\caption{Example webpage and natural image with elements representing their elements and text. The webpage one contains a random sample of 10 elements.}
\label{fig:elementsSample}
\end{figure}

In our work, we frame the problem of referring expression resolution as a segmentation problem. Our model outputs the probability of each pixel of the image being referred to, given a natural language referring expression. Unlike most previous work, ours focuses on images with labeled elements. Elements are often found in digital images, like the text and bounding boxes corresponding to the elements in the DOM (document object model\footnote {https://en.wikipedia.org/wiki/Document\_Object\_Model}) tree for a webpage image/snapshot, or the bounding boxes and text from OCR output of an image with text. Elements also can occur in natural images, for example an image with some objects annotated and labeled by image segmentation~\cite{DeepLab}. Figure~\ref{fig:elementsSample} demonstrates the elements of two images, a web page snapshot and a natural image from the COCO dataset. In the first example, the text associated with the elements are visibly present in the image, whereas in the second example, the text of the element is the label of the visual object covered by the bounding box. Note that because our proposed approach frames the referring expression resolution problem as segmentation, the model could also resolve a referring expression to part of the image that's not contained in an element. Text elements can be obtained from an OCR model run on the screen image, especially in the absence of accompanying meta-data. For example, visual text on a web page could be part of an integrated image and may not be included in the source html or DOM tree.

In this paper we present a novel, end-to-end trainable neural network architecture for referring expression resolution. The network inputs are the screen elements, user's natural language expression, and the image, as presented in Section 3. We show improved results over \cite{SegExp} where only the image is used and \cite{domnet} which uses a different architecture to process the elements. We present improvements with the proposed approach for resolving referring expressions on both webpages and natural images in Section 5.

See \url{http://www.rebrand.ly/refExpCode} for the code and data.

\section{Related Work}
Referring expression resolution on images has been studied in two research communities, conversational systems that respond back to users with visual information (possibly accompanied with system response utterances) and image processing. For conversational systems, the outlook of information users see are commonly designed by system builders~\cite{UiRefDilek} (for example, a list of movie entity cards displayed on the screen in response to user's request) or may be extracted from the corresponding meta-data (for example, DOM tree or html source of webpages shown to users), enabling extraction of image elements.
Hence we group the previous work on referring expression resolution into studies that work with natural images or webpage images with elements and natural images without elements.

\subsection{Resolving referring expressions on images with elements.}
Previous work investigated resolving referring expressions in user utterances of dialogue systems by using features that compute various forms of string matching features between user utterances and screen text or templates of positional expressions~\cite{UiRefDilek}. User's eye gaze~\cite{EyeGazeRefDilek} and pointing~\cite{PointingRefDilek} were also found as useful in addition to these textual features.
Most of these previous studies take elements into account as candidates and assume that the referring expression resolves to one of these screen elements. Due to this, most of these works also do not take the image into account. The task is modeled as estimating the probability of each element being the one referred to by the user's utterance, and usually the element with the highest probability is selected.

Other studies assume knowledge of organization of elements in the image and use the elements on the screen as context to help tag the referring expression with the parts that corresponds to the item name or position~\cite{UiRefAmazon}. 
The knowledge of image content ensures that when the referring expression is ``Click the third one", the element that is the third item on the screen is clear. However, in most web pages or natural images, such ordering may not be obvious. Our work is very different, since \cite{UiRefAmazon} frames the problem as simply tagging the referring expression, while ours segments the image.

Recent work on reinforcement learning (RL) on web interfaces using work flow-guided exploration~\cite{domnet} trains an agent to navigate simple webpages. They present a neural network architecture, Domnet, to process the elements on webpages, which we build off of in this work.
The architecture presented in World of Bits~\cite{worldOfBits} is unique since it uses the screen image, and outputs the probability the referring expression refers to each pixel location. This is the same output as our model. They also use features that compute if the referring expression matches the text of an element, but that part of the work is not clearly described for replication. Ours uses a more sophisticated method to processing the elements on the page and combine them with the image.

\subsection{Resolving referring expressions on natural images without elements.}
Earlier work on referring expressions from natural images assume that the candidates the expression could be referring to are known~\cite{RefMao}.
The first work to frame the referring expression resolution problem as a segmentation problem uses a convolutional neural network (CNN) to process the image and a recurrent neural network (RNN) to process a referring expression~\cite{SegExp}. It concatenates the results and uses another CNN to calculate the probability that each region is being referred to. We build upon this work in our paper.
This model was improved to process the referring expression word-by-word while accessing the image each time~\cite{SegExpRecLiu}. There are several papers that use a similar network to~\cite{SegExp} and apply reinforcement learning to train an agent to follow a natural language instruction~\cite{SegExpRlDm, SegExpRl2, SegExpRlFollowNet}. \cite{SegExpWeekSup} presents a network which produces referring expression segmentation masks in a weakly supervised way.

\section{Approach}

Our model combines information from the image, referring expression, and the screen elements. For every pixel of the image, the model outputs the probability of it belonging to the object being referred to.

We use an approach essentially the same as \cite{SegExp} to process the referring expression and image. An image segmentation model is used to process the image, and a text model encodes the referring expression. The text model (called $textmodel$ in the equation) can be any model which can encode text, for example a RNN or transformer network. The segmentation model is a fully convolutional neural network, so each of its intermediate layers have a width, height and depth. In order to combine other features with the segmentation model, we convert the features to have the same width and height. To do this for the referring expression, we tile the output of the text model to be the same size as an intermediate layer of the network by replicating it in the width and height dimensions. 

Let \(I\) be the image and \(R\) the referring expression. The equations for processing the image and referring expression are given by:

\begin{equation}
\begin{split}
R_{embed} &= textmodel(R) \\
R_{overlay} &= tile(R_{embed}) \\
I_{embed} &= CNN_1(I) \\
\end{split}
\end{equation}
\noindent where \(tile\)() replicates the embedding over height and width dimensions to create a 2 dimensional feature map.

Our work focuses on embedding the elements and combining them with the other relevant information. As in Domnet~\cite{domnet}, an embedding is calculated for each element on the page. However we calculate the embedding based on the text and the bounding box, where as Domnet~\cite{domnet} does not use a bounding box. We include the bounding box so expressions which refer to the items position can also be resolved. We embed the text with the same text model as for the referring expression, and concatenate the result with the normalized coordinates of the corresponding bounding box. This information is then fed through a few deep feed-forward neural network (DNN) layers.

If we let \(T_i\) and \(B_i\) be the text and the bounding box of the $i^{\rm th}$ element, respectively, then the embeddings of each element is given by:

\begin{equation}
E_i = DNN(concat(B_i, textmodel(T_i)))
\end{equation}
In Domnet~\cite{domnet}, attention is performed between each element and the referring expression. We also use attention in our approach, since it helps identify which elements are relevant to the referring expression. Attention is applied to the element embeddings with the following formulation:

\begin{equation}
\begin{split}
E_{i_{atten}} &= attention(R_{embed}, E_i) \\
attention(R_{embed}, E_i) &= \\ E_i*soft&max(DNN_a(R_{embed}) \cdot DNN_a(E_i)))
\end{split}
\end{equation}
\noindent where \(\cdot\) is the dot product, and * is an element wise multiplication. The softmax is applied over all of the elements. Note that the same network processes both input to attention, so the previous DNN must make $E_i$ the same size as $R_{embed}$.

In Domnet~\cite{domnet}, the embeddings for different elements are combined by averaging them. This can be applied to our network by averaging the embeddings and tiling the result so it can be combined with the image. The disadvantage of this approach, is that if there are many elements, it will be difficult for the network to preserve the relevant information of the elements. The network will also have to translate the information contained in the bounding box coordinates into the corresponding locations on the image.

\begin{figure*}[t]
\begin{center}
\includegraphics[width=.75\textwidth]{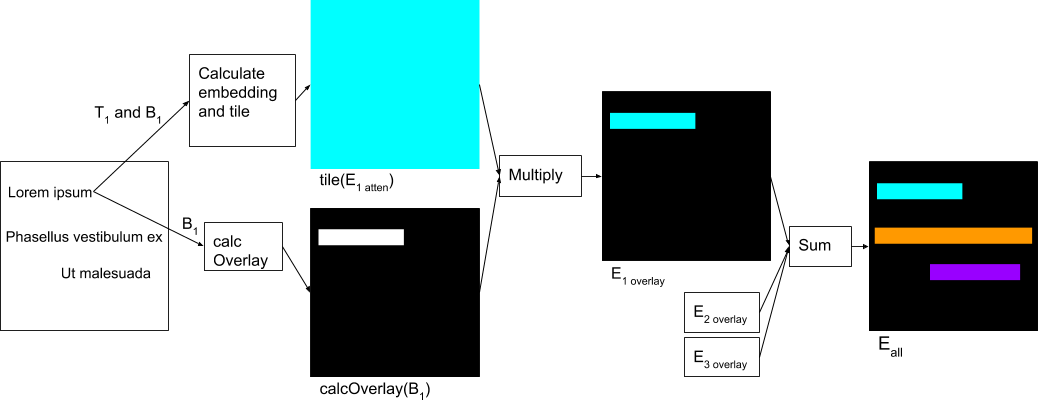}
\end{center}
\caption{Visualization of equation \ref{equ:proj} describing the projection and summing the overlays. The embedding dimension is 3 in this example, so it can be visualized as the red, green and blue channels of an image.}
\label{fig:project}
\end{figure*}

To overcome this disadvantage, we develop a novel method to combine the information of each element with the image information. We create a new feature map of the same size as the intermediate layer of the segmentation model. It is filled with the corresponding embedding values wherever one of the locations overlaps with an element bounding box, and is 0 everywhere else. If a location overlaps with multiple bounding boxes, the corresponding embeddings are summed together. This way, less of the elements have to be averaged together, and the rest of the segmentation model can easily localize the element. This method is calculated by:

\begin{equation}
\begin{split}
E_{i_{overlay}} = calcOverlay(B_i) * tile(E_{i_{atten}}) \\
E_{all} = E_{1_{overlay}} + E_{2_{overlay}} + ... E_{N_{overlay}}
\end{split}
\label{equ:proj}
\end{equation}
\noindent
where \(calcOverlay\) generates a 3-dimensional feature map with 1s where the coordinates are overlayed by the bounding box and 0s everywhere else. $E_{i_{overlay}}$ is a feature map with element i projected onto its corresponding bounding box. See figure~\ref{fig:project} for a visualization of this method. This feature map is also put through a small CNN for further processing.

At this point, there is a feature map for the image, referring expression and the elements. Since the image feature map is processed by a CNN, its depth is greater than the depth of the image, and its width and height are smaller. We combine the feature maps by concatenating them in the depth dimension, and feeding the result through a small CNN. We then sum the result with the original image feature map similar to a residual connection. Summing the result with the original feature map allows us to use a pre-trained network. After the combined feature map is calculated in this way, a CNN is used to process it. The softmax function is used to convert this to probabilities for each pixel. This is calculated as follows:

\begin{equation}
\begin{split}
F = CNN_3(concat(R_{overlay}, I_{embed}, CNN_2(E_{all}))) \\
seg = softmax(CNN_4(I_{embed}+F))
\end{split}
\end{equation}

\begin{figure}[h]
\begin{center} 
\includegraphics[width=.42\textwidth]{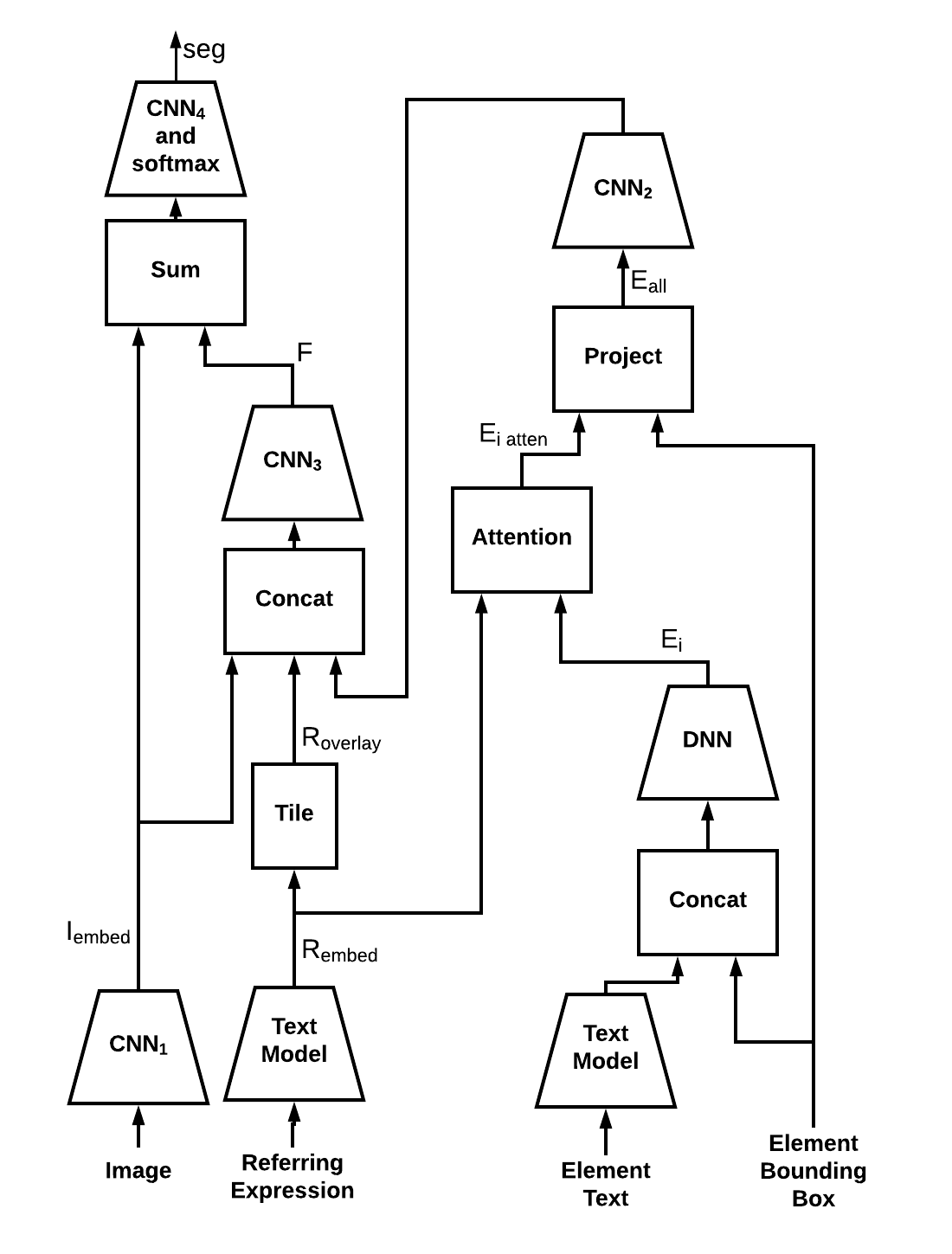}
\end{center}
\caption{Model architecture.}
\label{fig:ElementsDiagram}
\end{figure}

Figure~\ref{fig:ElementsDiagram} shows the overall architecture used in our work with inputs and outputs.

\section{Implementation details}
We use the DeepLab~\cite{DeepLab} architecture for segmentation, with MobileNetV2~\cite{MobileNet} as the backbone. We use a byte level Universal Sentence Encoder~\cite{TextEnc} to encode the text from each element and the referring expression. To prevent over-fitting we use a pre-trained Universal Sentence Encoder~\cite{TextEnc} and don't fine tune it, however fine tuning is possible with our architecture. The DeepLab~\cite{DeepLab} we use is pretrained on COCO segmentation and fine tuned during training.

\section{Experiments and Results}

\subsection{Evaluation Metrics}
We use mean intersection over union (mIOU) since that is a standard segmentation metric used in previous work. IOU is defined as the area that the predicted segmentation intersects with the ground truth, divided by the area of their union. We also measure the percentage of the time the pixel that the model predicts as the highest probability is classified correctly. We call this accuracy. It measures the percentage of the time an agent would click on the correct item if it clicked on the highest probability pixel. For our use case, the accuracy metric is the most important, as it could be used to identify the object that the natural language expression is referring to.

\subsection{Coco Dataset}
\subsubsection{Dataset Description}
We experiment on the refcocog~\cite{refcocog}, refcoco~\cite{uncDatasets} and refcoco+~\cite{uncDatasets} datasets built off of COCO \cite{coco}. We use the ground truth bounding boxes and class labels as the elements for this dataset. We use the text of the class label for the text of each element. We use the same train validation and test split as in~\cite{datasetSplits} and for the refcoco and refcoco+ we combine test set A and B in our test set. 

\subsubsection{Results}
\begin{table*}[h]
\centering
\begin{tabular}{|l|l|l|l|l|l|l|l|l|l|}
\hline
Img & El & Proj & Description             & \multicolumn{2}{c|}{refcoco+}        & \multicolumn{2}{c|}{refcoco}       & \multicolumn{2}{c|}{refcoco google}     \\ \hline
    &    &      &                                     & Accuracy & mIOU     & Accuracy & mIOU   & Accuracy & mIOU   \\ \hline
1   & 1  & 1    & Ours                                & 0.6466   & 0.4929   & 0.8115   & 0.6375 & 0.6808   & 0.5005 \\ \hline
1   & 0  & 0    & Image only \cite{SegExp}            & 0.573    & 0.3863   & 0.7393   & 0.5219 & 0.5783   & 0.3782 \\ \hline
0   & 1  & 1    & Ours elements only                  & 0.44     & 0.1575   & 0.6741   & 0.3644 & 0.52     & 0.2237 \\ \hline
1   & 1  & 0    & Ours without projection             & 0.5715   & 0.3829   & 0.7488   & 0.531  & 0.5733   & 0.373  \\ \hline
0   & 1  & 0    & No proj EL only like \cite{domnet}  & 0.3911   & 0.1755   & 0.6015   & 0.309  & 0.4547   & 0.1876 \\ \hline
\end{tabular}
\caption{Results of different methods on coco datasets. 1 and 0 are used to denote that feature was or was not presented to the model.}
\label{tab:coco}
\end{table*}

\begin{table*}[t]
\centering
\begin{tabular}{|l|l|l|l|l|l|l|}
\hline
Img & El & DomNet Features & Proj & Description                  & Accuracy & mIOU    \\ \hline
1   & 1  & 0               & 1    & Ours                         & 0.4049   & 0.22    \\ \hline
1   & 1  & 1               & 1    & Ours + DomNet\cite{domnet} features       & 0.4153   & 0.2227  \\ \hline
1   & 0  & 0               & 0    & Image only \cite{SegExp}                  & 0.2684   & 0.1353  \\ \hline
1   & 1  & 1               & 0    & DomNet\cite{domnet} elements method + Img & 0.2507   & 0.1238  \\ \hline
0   & 1  & 0               & 1    & Ours elements only           & 0.2815   & 0.1116  \\ \hline
0   & 1  & 1               & 0    & DomNet\cite{domnet} elements method       & 0.0582   & 0.00981 \\ \hline
\end{tabular}
\caption{Results of different methods on our website referring expression dataset.}
\label{tab:WebRef}
\end{table*}

We tested our re-implementation of the model from~\cite{SegExp} on the ReferIt dataset~\cite{uncDatasets}, and found that our re-implementation achieves the same mIOU to theirs.

We experimented with the following variations in the model:
\begin{itemize}
  \item Img: Whether or not the image was given to the model. 
  \item El: Whether or not to process the elements of the model. If false, the elements embeddings aren't concatenated with the image and referring expression features.
  \item Proj: If true uses the projection method described. If false, tiles the average of each elements embedding to the same size of the image instead of projecting.
\end{itemize}

The model hyperparameters are the same in every experiment and were optimized on the model from~\cite{SegExp} on the refcocog~\cite{refcocog} dataset. We used early stopping on the "Ours without img" runs.

The results of our experiments are shown in Table~\ref{tab:coco}.
From these results, we conclude the following: With our projection method, access to the elements helps over only using the image in terms of both evaluation metrics: mIOU and accuracy. However, when the tile method is used instead, the elements do not help. The methods which are given only the elements and not the image do poorly. On most datasets and evaluation measures, the projection method does better than the one using the tile method when the model is only given the elements.

We don't attempt to show improvement over the state of the art on the refcoco datasets. Rather, we show that a model with information about the elements does better than a model with only the image, and that our projection method does better than averaging elements as in~\cite{domnet}.

The first example, presented in the top row of Figure~\ref{fig:results} shows the advantage of using both the image and elements on the ref coco dataset. The model with only the image segmented part of the incorrect zebra. The model with only the elements segmented the correct region, but it couldn't identify the shape of the zebra. The model with the elements and image was able to identify the shape of the correct zebra, and use the element bounding box to ignore the incorrect one.

\subsection{Webpage Dataset}
\subsubsection{Dataset Description}
\begin{figure*}[h]
    \centering
    \begin{subfigure}[h]{\textwidth}
        \begin{subfigure}[b]{0.19\textwidth}
            \includegraphics[width=\textwidth]{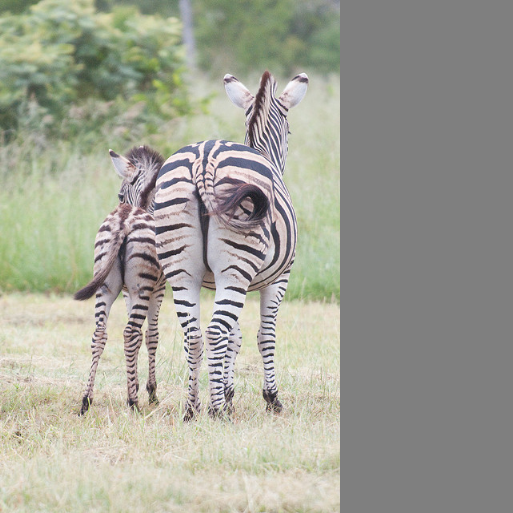}
        \end{subfigure}
        \begin{subfigure}[b]{0.19\textwidth}
            \includegraphics[width=\textwidth]{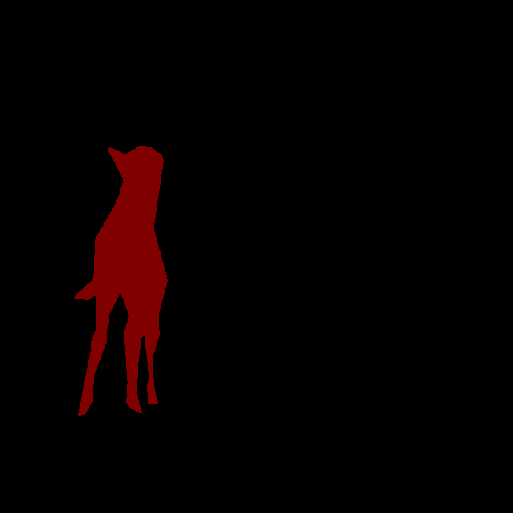}
        \end{subfigure}
        \begin{subfigure}[b]{0.19\textwidth}
            \includegraphics[width=\textwidth]{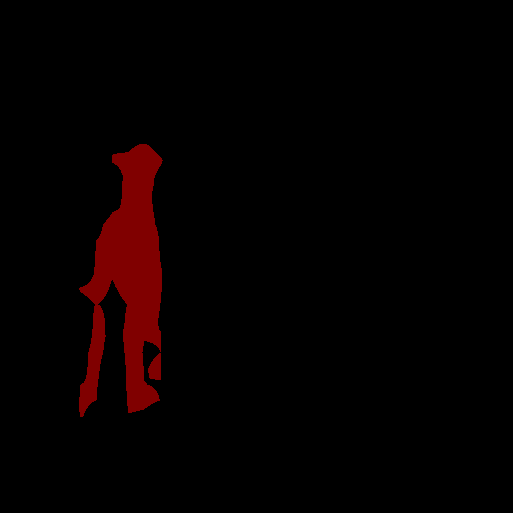}
        \end{subfigure}
        \begin{subfigure}[b]{0.19\textwidth}
            \includegraphics[width=\textwidth]{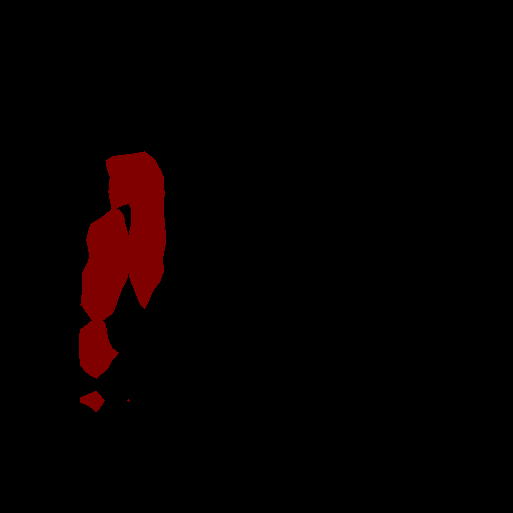}
        \end{subfigure}
        \begin{subfigure}[b]{0.19\textwidth}
            \includegraphics[width=\textwidth]{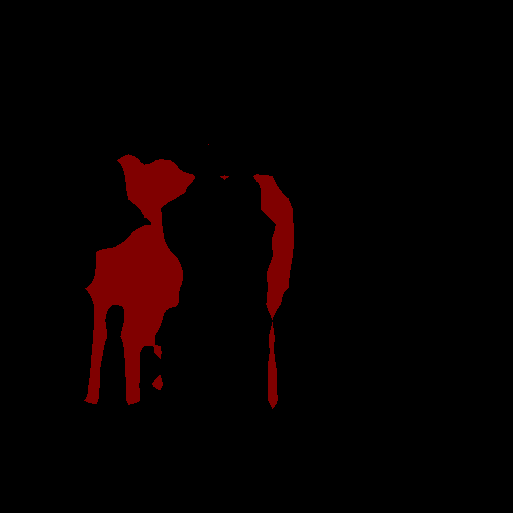}
        \end{subfigure}
        \captionsetup{labelformat=empty}
        \caption{Referring Expression: smaller one}
    \end{subfigure}
    \begin{subfigure}[b]{\textwidth}
        \begin{subfigure}[b]{0.19\textwidth}
            \includegraphics[width=\textwidth]{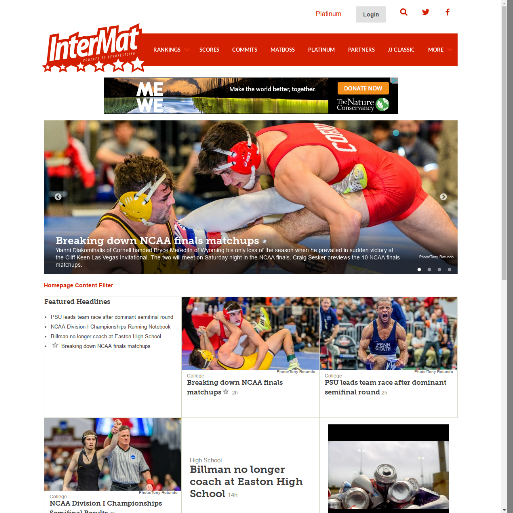}
            \captionsetup{labelformat=empty}
            \caption{Image}
        \end{subfigure}
        \begin{subfigure}[b]{0.19\textwidth}
            \includegraphics[width=\textwidth]{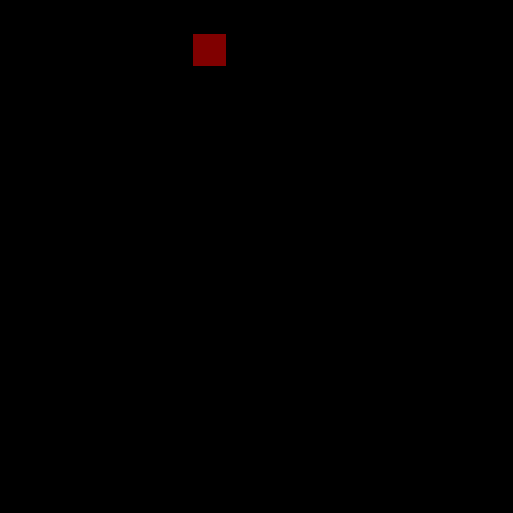}
            \captionsetup{labelformat=empty}
            \caption{Ground truth}
        \end{subfigure}
        \begin{subfigure}[b]{0.19\textwidth}
            \includegraphics[width=\textwidth]{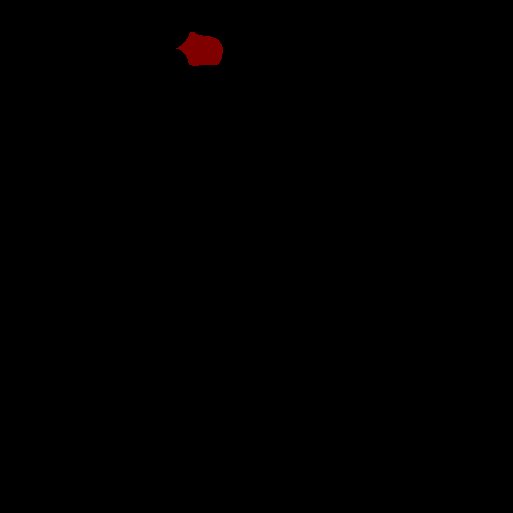}
            \captionsetup{labelformat=empty}
            \caption{Ours}
        \end{subfigure}
        \begin{subfigure}[b]{0.19\textwidth}
            \includegraphics[width=\textwidth]{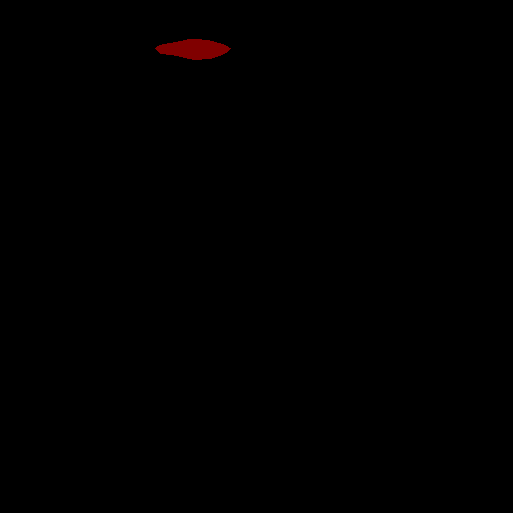}
            \captionsetup{labelformat=empty}
            \caption{Ours elements only}
        \end{subfigure}
        \begin{subfigure}[b]{0.19\textwidth}
            \includegraphics[width=\textwidth]{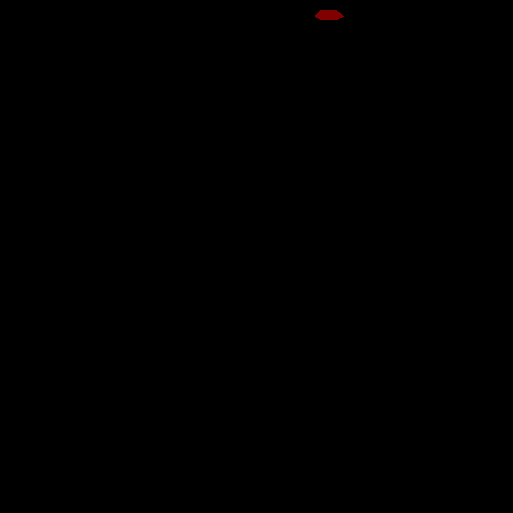}
            \captionsetup{labelformat=empty}
            \caption{Image only}
        \end{subfigure}
        \captionsetup{labelformat=empty}
        \caption{Referring Expression: option left to commits in the red bar}
    \end{subfigure}
    
    \caption{Results of different methods on the refcoco+ dataset and our webpage dataset. The method names correspond to those in the results tables. These examples are chosen to illustrate the differences between the methods. The segmentation masks are obtained by thresholding the probability predicted by the model at .5.}
    \label{fig:results}
\end{figure*}

We also experiment on web page images since this is an obvious use case for this approach. To our knowledge, there is no existing public dataset for resolving referring expressions on web page images. So we used a service similar to Mechanical Turk to collect our own dataset. We will release this dataset upon publication. We collected about 104k image and referring expression pairs. There are about 10 referring expressions per webpage and most referring expressions refer to a different user interface item of the image. In our dataset, the elements are groups of words identified by an OCR engine, with bounding boxes also output by OCR.

For each webpage, we used heuristics on the dom tree to identify a webpage element a user can refer to in their utterance, like buttons and text fields. The area within the bounding box of the element gives the ground truth segmentation. We sent the images with the bounding box of the targeted element highlighted to crowd workers and they typed one utterance that includes a referring expression, for each image and element pair.

To obtain the elements for input to the model, we used an optical character recognition engine to identify groups of words on the screen. The text and bounding boxes of those groups are fed into the model as the elements. Note that our model will work the same with elements from other sources, such as the dom tree.

The dataset has diverse referring expressions. Some refer to the text directly, for example ``click the login button". The rest refer to the elements
relative to each other, by their position on the screen, or color or content. For example ``the green one" or ``the third one in the menu bar".

We divide the dataset into training validation and test by image, so that the same image does not appear in multiple splits. We use about 10\% for validation and test and the remaining for training. All results shown here are on the test split. To simulate a dialogue, we convert all referring expressions to lowercase and remove special characters.

\subsubsection{Results}
In order to compare against the methods to combine elements from Domnet~\cite{domnet} we implement the techniques they use for encoding the elements. In Domnet~\cite{domnet}, the embeddings of the element text that matches the referring expression is part of the embedding for each element. The embeddings of the nearby elements are also summed together and added as part of the embedding.

We experiment with the variations defined in the coco experiments for whether or not to use the image, the elements or the projection method. We also experiment with additional model variations that control whether the features from Domnet~\cite{domnet} are used, shown by the ``DomNet Features" column in table \ref{tab:WebRef}. We used the same hyperparameters in every configuration that we optimized on the image only model. We also applied some dropout to the elements processing model which we tuned on the elements only model.

Our results in table \ref{tab:WebRef} show that using the image and our elements and projection method does significantly better than using the image or elements by themselves. The method using only the elements works about the same as the method using only the image. Using the elements with the tile method and the image does not do any better than only using the image. Using the method to combine the elements presented in DomNet\cite{domnet} by using the Domnet features and the tile method without the image does poorly. Also, the element combining method from DomNet~\cite{domnet} with the image does not do any better than only using the image. This is probably because the DomNet method was designed to select a candidate and not to produce a segmentation output. Our results show that simply using the method from DomNet~\cite{domnet} to combine the elements does not work well for our problem. The features used in Domnet~\cite{domnet} do not give any improvement over only using the text and bounding boxes in each element.

These results support the conclusion that using the elements performs better than using only the image, and using our projection method works better than the element processing method used by Domnet~\cite{domnet}.

The example in the bottom row of Figure \ref{fig:results} shows the advantage of using both the image and elements in the webpage task on an example image and referring expression. The model with only the image identified the wrong location since it did not know how to read the text on the page. The model with only the elements could segment out the element to the left of the ``commits" element, but it did not make the identified region the correct size and shape. The model with the elements and image could identify the correct region, and could deduce the correct size and shape from the image.

\section{Conclusions}
We presented a referring expression resolution approach that inputs image, image elements represented by corresponding text and bonding box of each element and a natural language referring expression, and outputs the probability for each pixel belonging to the object referred to in the natural language expression. In experimental results on two datasets, we demonstrated that our method of projecting the element embeddings onto a feature map works better than previous methods from DomNet\cite{domnet} for incorporating the elements. We also demonstrated that using both the image and elements works better than using either on its own.

In this work we used the elements information to aid in the task of resolving referring expressions, however it could be used to aid in other tasks, such as visual question answering with a webpage image or semantic segmentation of an image with text.

\bibliographystyle{IEEEbib}
\bibliography{strings,refs}

\end{document}